\documentclass[sigconf,natbib=false]{acmart}

\AtBeginDocument{%
  }

\setcopyright{acmlicensed}
\copyrightyear{ }
\acmYear{ }
\acmDOI{XXXXXXX.XXXXXXX}
\usepackage[backend=bibtex,style=numeric]{biblatex}
\acmConference[Conference acronym 'XX]{Make sure to enter the correct
  conference title from your rights confirmation emai}{June 03--05,
   }{Woodstock, NY}
\acmISBN{ }

\begin{document}

\title{Optimizing Autonomous Driving for Safety: A Human-Centric Approach with LLM-Enhanced RLHF}


\author{ Yuan Sun }
\affiliation{%
  \institution{WINLAB, Rutgers University}
  \city{ Piscataway, NJ, }
  \country{ USA}}
  \email{ ys820@soe.rutgers.edu }

\author{ Salami Pargoo, Navid }
\affiliation{%
  \institution{WINLAB, Rutgers University}
  \city{ Piscataway, NJ, }
  \country{ USA}}
  \email{ navid.salamipargoo@rutgers.edu }

\author{ Peter J. Jin }
\affiliation{%
  \institution{Rutgers University}
  \city{ Piscataway, NJ, }
  \country{ USA}}
  \email{ peter.j.jin@rutgers.edu }

  \author{ Jorge Ortiz }
\affiliation{%
  \institution{WINLAB, Rutgers University}
  \city{ Piscataway, NJ, }
  \country{ USA}}
  \email{ jorge.ortiz@rutgers.edu }

\renewcommand{\shortauthors}{  }

\begin{abstract}

Reinforcement Learning from Human Feedback (RLHF) is popular in large language models (LLMs), whereas traditional Reinforcement Learning (RL) often falls short. Current autonomous driving methods typically utilize either human feedback in machine learning, including RL, or LLMs. Most feedback guides the car agent's learning process (e.g., controlling the car). RLHF is usually applied in the fine-tuning step, requiring direct human "preferences," which are not commonly used in optimizing autonomous driving models. In this research, we innovatively combine RLHF and LLMs to enhance autonomous driving safety. Training a model with human guidance from scratch is inefficient. Our framework starts with a pre-trained autonomous car agent model and implements multiple human-controlled agents, such as cars and pedestrians, to simulate real-life road environments. The autonomous car model is not directly controlled by humans. We integrate both physical and physiological feedback to fine-tune the model, optimizing this process using LLMs. This multi-agent interactive environment ensures safe, realistic interactions before real-world application. Finally, we will validate our model using data gathered from real-life testbeds located in New Jersey and New York City.

\end{abstract}





\maketitle

\section{Introduction}

Reinforcement Learning (RL) and Large Language Models (LLMs) play a crucial role in the development of autonomous driving systems. RL, a branch of machine learning, focuses on enabling agents to make optimal decisions over time by learning from their mistakes and experiences. Numerous studies have demonstrated the application of RL in autonomous driving. For instance, \cite{galias2019simulation} proposed using RL to map sensor observations to control outputs in simulated environments. Similarly, \cite{lillicrap2015continuous} explored deep RL for continuous control tasks, which are applicable to autonomous driving scenarios. Additionally, \cite{wang2023efficient} introduced ASAP-RL, an efficient RL algorithm that utilizes motion skills and expert priors to enhance learning efficiency and driving performance in dense traffic. Moreover, \cite{hoel2018automated} presented a method for making decisions such as lane changing, accelerating, and braking on highways, employing Deep Q Networks (DQN) to train their model and predict optimal actions. Recent research has increasingly incorporated Large Language Models (LLMs) into autonomous driving systems, leveraging their capabilities for decision-making, reasoning, and interaction. For instance, \cite{cui2023drivellm,fu2024limsim++,rayfeedback} integrated LLMs to enhance commonsense reasoning and high-level decision-making. In another study, \cite{duan2024prompting} utilized LLMs to help autonomous driving models mimic human behavior, thereby improving end-to-end driving performance. \cite{wang2022adept} employed GPT to extract crucial information from NHTSA accident reports using a QA approach, enabling the generation of diverse scene codes for simulation and testing. Additionally, \cite{tan2023language} demonstrated the use of LLMs as powerful interpreters that translate user text queries into structured specifications of map lanes and vehicle locations for traffic simulation scenarios.
RLHF is a fundamental component in the training of Large Language Models (LLMs) and is regarded as an essential element of the modern LLM training pipeline \cite{kirk2023understanding, zheng2024balancing, zhu2023principled}. RLHF is particularly well-suited for LLMs \cite{sun2023reinforcement}, as it involves RL agents learning from human preference feedback. This type of feedback is considered more intuitive for human users, better aligned with human values, and easier to obtain in various applications \cite{kirk2023understanding}.

However, most of the work applies RLHF to optimize LLMs because the scenarios in which humans use LLMs are more suitable for tracking human preferences. This alignment with human values ensures that the models perform more intuitively in real-world applications. In contrast, for autonomous driving scenarios, it is impractical for humans to provide preference feedback on a frame-by-frame basis. Consequently, RLHF is seldom used in autonomous driving contexts.

In this research, we creatively apply RLHF to autonomous driving by modeling human preferences through various sensor feedback from our environment. We incorporate both physical and physiological feedback into the simulation to optimize the RL training loop for autonomous driving. Additionally, the LLM agent facilitates interaction within our multi-agent system. We posit that training autonomous car agents alongside human-driven cars in the simulation can significantly enhance safety and allow the agents to learn human preferences concurrently. This approach aligns the autonomous car models more closely with real-world scenarios, ultimately making the application of autonomous driving models safer in real-life contexts.

\section{Methodology}

\begin{figure*}[h]
  \centering
  \includegraphics[width=0.8\linewidth]{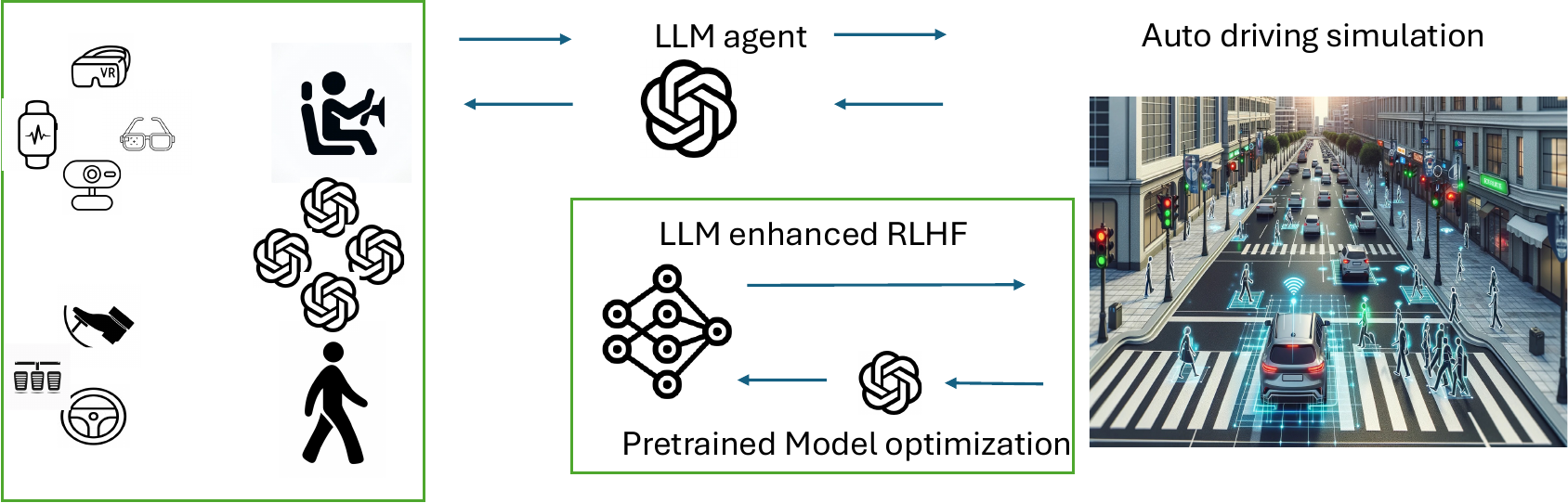}
\caption{Overview of our human-centric multi-agent LLM-enhanced RLHF system framework. During the fine-tuning of an autonomous car model, human agents and proliferated LLM agents mimicking multiple human behaviors are incorporated into the environment to align with real-world human preferences.}
  \label{overview}
\end{figure*}

Our system (figure~\ref{overview}) is designed as a multi-agent framework, centering on human interaction while incorporating LLM agents and autonomous car agents. The system includes human drivers, human pedestrians, and an LLM agent that mimics their behavior to generate training interactions for the car. Humans control the car using physical controllers such as steering wheels and pedals. Additionally, they wear various wearable sensors, including VR headsets, wristbands to collect and monitor physiological signals in real-time, and smart glasses to track focus. A camera is also set up in the environment to record human reactions. This collected multimodal data is sent to the simulation. Before integration, the LLM assists the car agent in understanding this data and aids the human agent in adapting to the simulation environment. The autonomous vehicle model learns from human feedback within the RL loop, with the LLM helping to interpret human data into "preferences" to optimize the model in the RL loop.

\subsection{RLHF}

In traditional reinforcement learning, the objective of the agent is to develop a policy, a function that dictates its actions. This policy is optimized to maximize rewards provided by a distinct reward function based on the agent's performance in a given task \cite{russell2016artificial}. However, defining a reward function that accurately reflects human preferences is challenging. To address this, RLHF aims to train a "reward model" directly from human feedback \cite{ziegler2019fine}.

However, in our scenario, it is typically difficult to obtain "preference" feedback directly on a frame-by-frame basis. For instance, when training an autonomous car model using RL, the car might simply reward itself with a positive score for avoiding collisions. However, the car might execute a rapid lane change that frightens the user. This type of feedback—reflecting user comfort and safety preferences—is crucial but challenging to capture. Additionally, scenarios such as aggressive braking, abrupt acceleration, or failing to yield to pedestrians can all contribute to negative user experiences, which are important "preferences" in the autonomous driving RL loop.At the same time, human driving behavior provides valuable feedback to the system, as it reflects real-world driving preferences and responses to various driving conditions.

In our autonomous driving RLHF framework, the objective function\cite{stiennon2020learning} is defined as follows:

\begin{equation}
\text{objective}(\phi) = \mathbb{E}_{(x,y) \sim D_{\pi^{RL}_{\phi}}} \left[ r_{\theta}(x,y) - \beta \log \left( \frac{\pi^{RL}_{\phi}(y|x)}{\pi^{SFT}(y|x)} \right) \right] 
\end{equation}

where ,in our work, \( x \) represents the input data from various sensors, including physical sensors, physiological sensors, and simulation data such as LiDAR and camera inputs. The output \( y \) denotes the actions taken by the autonomous vehicle. The reward function \( r_{\theta}(x,y) \) evaluates the quality of the action \( y \) given the sensor inputs \( x \), guided by human feedback. The term \( \beta \log \left( \frac{\pi^{RL}_{\phi}(y|x)}{\pi^{SFT}(y|x)} \right) \) introduces a KL divergence penalty to ensure the new policy remains close to the initial supervised model, balancing learning from new data while retaining useful information from the initial model.

\subsection{LLM Integration in Autonomous Driving}

The primary functions of the LLM in our work include acting as an agent in the simulation, facilitating interaction between humans and the simulation system, and optimizing the RL training loop.

\subsubsection{LLM Agent in the Simulation}

In this section, we emphasize our multi-agent system with two key use cases. First, the LLM agent can mimic human behavior to interact with the car agent when a human is not available. Second, when a human is in the loop, the LLM agent can act as another agent, such as a car or pedestrian, to increase the system's complexity. For instance, human feedback differs when there is one car versus multiple cars on the road. The feedback in such a complex scenario is more representative of real-life situations.

\subsubsection{Enhanced Human-Simulation Interaction Based on LLM}

When the simulation interacts with the human agent, the LLM can enhance the interaction. Firstly, when sending data collected from humans to the system, the LLM can help interpret the data. For example, if a driver is skilled, the LLM might adjust the simulation weather to foggy conditions. Conversely, if the driver is less skilled, the LLM can help the user adapt to the environment before the car agent begins its training.

\subsubsection{LLM-Enhanced RLHF}

In the RLHF loop, "preferences" are not as straightforward as yes or no answers. The LLM can translate physical and physiological data into preference formats, which are then incorporated into the objective function. For example, if a driver's heart rate increases significantly during a particular maneuver, the LLM can interpret this physiological response as a negative preference for that action. Similarly, if sensor data indicates smooth and confident handling of the vehicle, this can be translated into a positive preference. By integrating these nuanced preferences into the RL training loop, the autonomous driving model can be better aligned with human comfort and safety standards.

\section{hardware setup}

\begin{figure}[h]
  \centering
  \includegraphics[width=\linewidth]{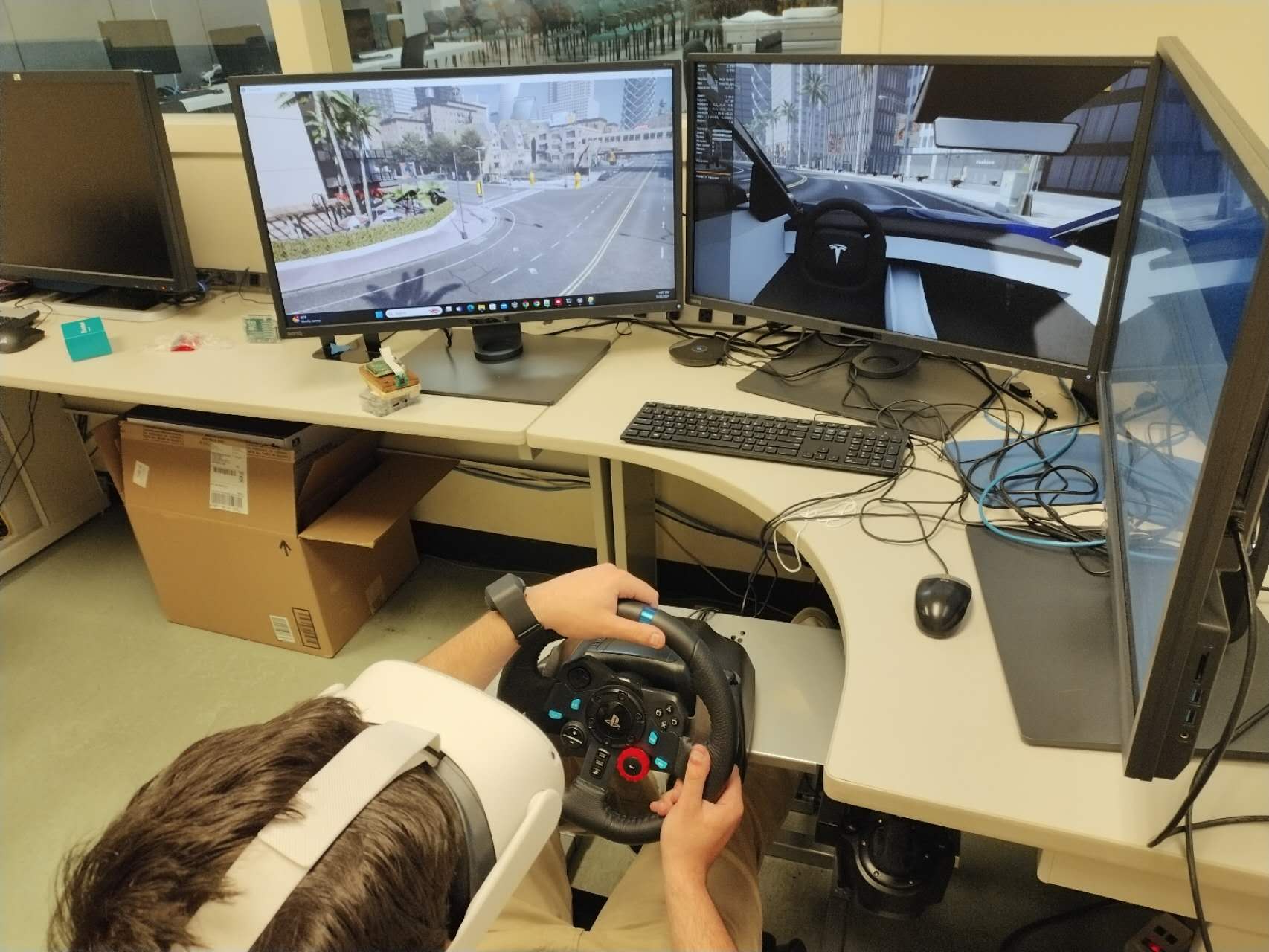}
  \caption{Simulation room with VR headset, steering controls, and monitors for real-time multimodal data collection and autonomous driving optimization.}
  \label{overview}
\end{figure}

The sensors utilized in our multi-agent LLM-enhanced RLHF system are categorized into two primary modalities: vehicle and physiological. The vehicle sensors, primarily sourced from the CARLA simulator and Logitech hardware, include acceleration, rotation (gyroscope), speed, brake, steering, throttle, and reverse, all sampled at approximately 60 Hz. These sensors capture the dynamic state and control inputs of the autonomous vehicle. The physiological sensors, provided by Empatica, measure various physiological signals such as blood volume pulse (64 Hz), heart rate (1 Hz), interbeat interval (varying), electrodermal activity (4 Hz), wrist acceleration (32 Hz), and body temperature (4 Hz). Additionally, the gaze modality employs an Adhawk sensor to track coordinates on the screen at a sampling rate of 125 Hz, capturing the human agent’s visual focus and attention. These sensors monitor the human agent’s physical and emotional responses during the simulation. Except for the monitor, we also have a VR headset to create an immersive environment. Additionally, a Raspberry Pi camera in the simulation room observes the human reaction to the simulation. This multi-modal data integration is crucial for fine-tuning the autonomous driving model, providing comprehensive feedback to align the model’s performance with human preferences and ensuring realistic and safe interactions in the simulation environment.

\section{Initial Implementation}

Our initial implementation demonstrates the integration of the LLM with the car simulation system using the GPT-4 interface. The LLM agent can imitate human driving behavior, especially when interacting with a car agent in front. It also assists the car agent in managing situations such as avoiding collisions. Additionally, the LLM agent helps human users by instructing them on how to effectively navigate and use the simulation system.

\begin{figure}[h]
  \centering
  \includegraphics[width=\linewidth]{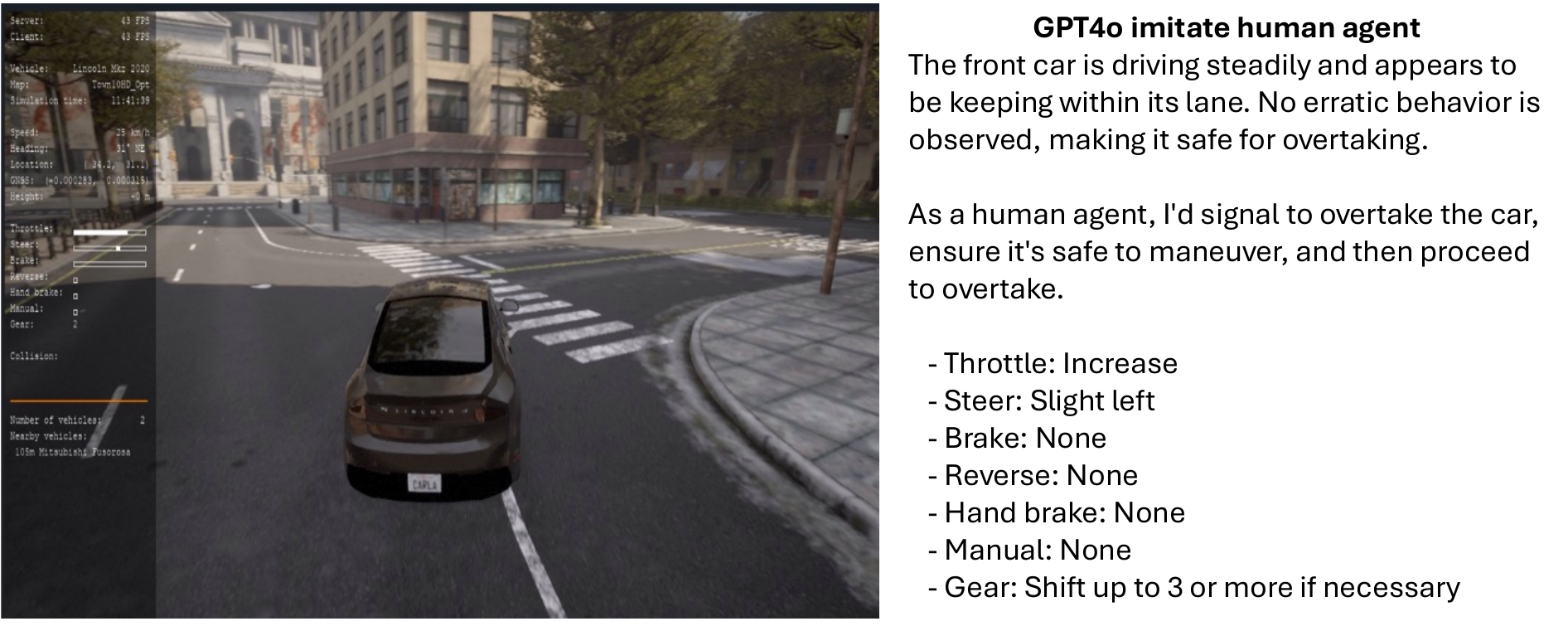}
\caption{Example of GPT-4o imitating a human agent in the CARLA simulation. The LLM agent is attempting to overtake the car in front in a human-like manner.}
  \label{imitate}
\end{figure}

\begin{figure}[h]
  \centering
  \includegraphics[width=\linewidth]{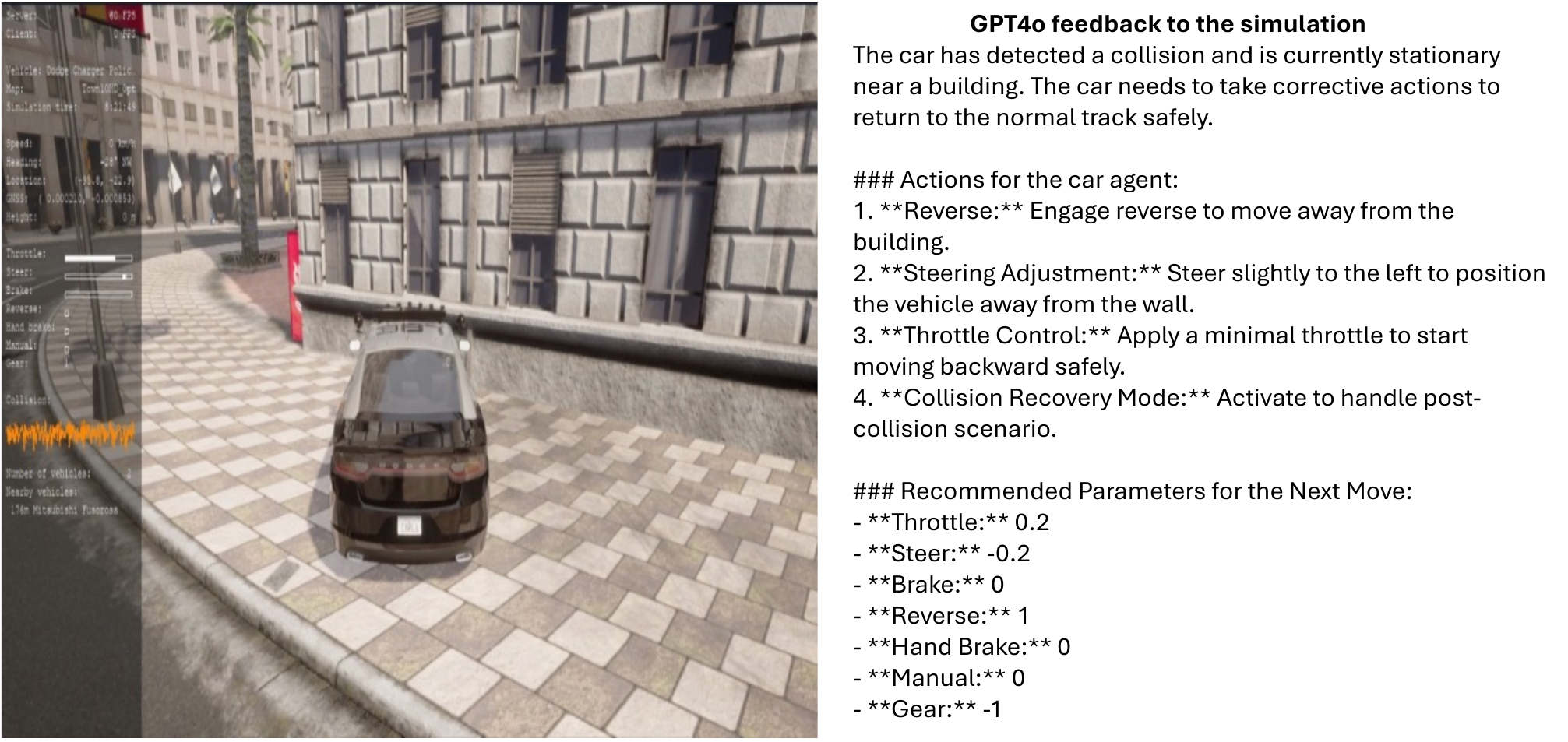}
  \caption{Example of the LLM agent guiding the autonomous car agent to reverse away from a collision with a building.}
  \label{feedbacktosys}
\end{figure}

\begin{figure}[h]
  \centering
  \includegraphics[width=\linewidth]{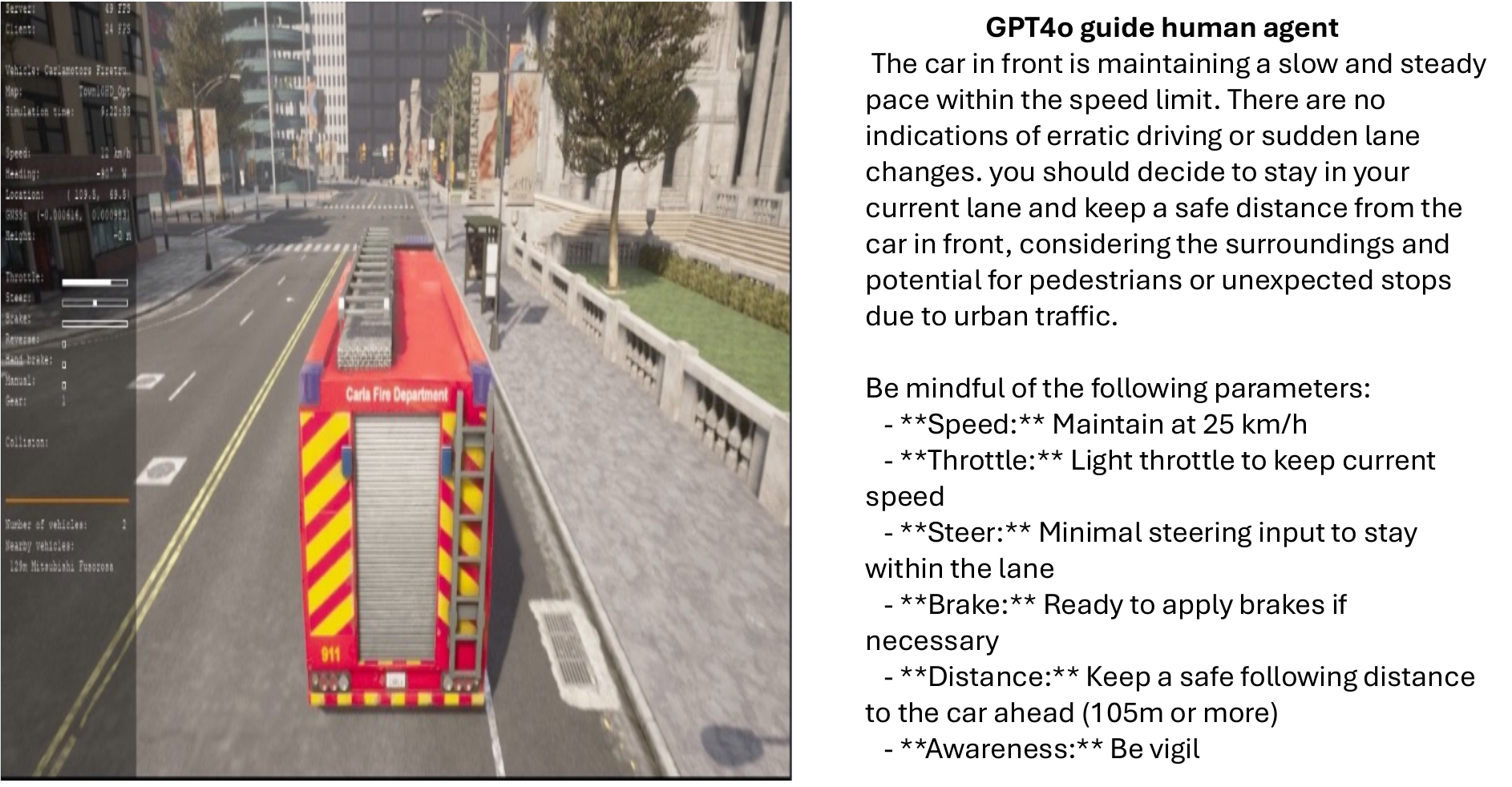}
 \caption{Example of the LLM agent assisting the human agent in using the simulation.}
  \label{helphuman}
\end{figure}

Furthermore, we plan to implement our experiments in our real-life city test bed, located in Harlem\cite{cosmos-lab}, NYC, and New Brunswick\cite{cait-datacity}, NJ. The real-life data collected from these locations will be used to test the robustness of our algorithm. Additionally, we can import real-life data into the simulation as a cross-validation method. The figure~\ref{NB_c} below shows an example of importing real-life road data from New Brunswick into the CARLA system.

\begin{figure}[h]
  \centering
  \includegraphics[width=0.5\linewidth]{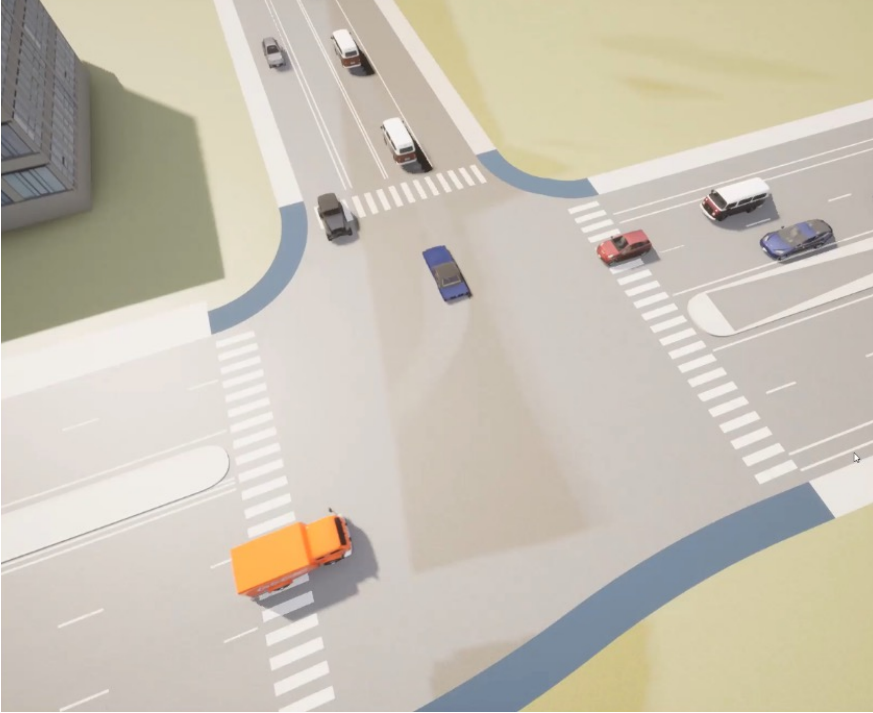}
\caption{Example of importing real-life road data from New Brunswick testbed into the CARLA simulation.}
  \label{NB_c}
\end{figure}

\section{Conclusion and Future Work}

In this work, we introduce a novel framework that integrates RLHF and LLMs to optimize autonomous driving models. We define human preferences within the RLHF framework and build a simulation-to-reality system based on this concept. Our method simulates a multi-agent environment for training the car agent, allowing it to learn human behaviors through multimodal sensory data. The LLM agent can proliferate multiple human agents by mimicking human behavior and facilitating interactions between the car agent and other agents on the road within the simulation. When optimizing the model, the LLM agent also interprets human data to enhance the model via RLHF. Our system incorporates both physical and simulation sensors. The initial implementation demonstrates various scenarios where LLMs are applied to the framework. This preliminary work establishes the foundational infrastructure for our experiments and discusses the theoretical feasibility of the framework.

However, there is still much work to be done in the next stage of our plan. Firstly, the GPT-4 interface has rate limits; we may need to explore different interfaces for this study. The machine learning model for autonomous driving should be evaluated across different types of multimodal models to prove the robustness of our method. We will apply more real-life data to our research to improve the robustness of our method. We will recruit subjects with diverse backgrounds and varying driving skills for human evaluation. Individuals with good driving skills and those with less experience present different challenges in our study. We plan to provide a comprehensive evaluation of how different levels of background influence the RLHF autonomous driving framework.

Finally, we hope that through our study, we can eventually propose a safe driving model that can help autonomous vehicles navigate real-life roads and contribute to the overall road safety of society.

\section{Acknowledgments}

This work was supported by the National Science Foundation (NSF) as part of the Center for Smart Streetscapes, under NSF Cooperative Agreement EEC-2133516. DataCity Smart Mobility Testing Ground is jointly funded by Middlesex County Resolution 21-821-R, New Jersey Department of Transportation and Federal Highway Administration Research Project 21-60168.

\printbibliography

@String{Chelsea = "Chelsea" }

@article{galias2019simulation,
  title={Simulation-based reinforcement learning for autonomous driving},
  author={Galias, Christopher and Jakubowski, Adam and Michalewski, Henryk and Osi{\'n}ski, B{\l}a{\.z}ej and Zi{\k{e}}cina, Pawe{\l}},
  year={2019}
}

@article{lillicrap2015continuous,
  title={Continuous control with deep reinforcement learning},
  author={Lillicrap, Timothy P and Hunt, Jonathan J and Pritzel, Alexander and Heess, Nicolas and Erez, Tom and Tassa, Yuval and Silver, David and Wierstra, Daan},
  journal={arXiv preprint arXiv:1509.02971},
  year={2015}
}

@article{wang2023efficient,
  title={Efficient reinforcement learning for autonomous driving with parameterized skills and priors},
  author={Wang, Letian and Liu, Jie and Shao, Hao and Wang, Wenshuo and Chen, Ruobing and Liu, Yu and Waslander, Steven L},
  journal={arXiv preprint arXiv:2305.04412},
  year={2023}
}

@inproceedings{hoel2018automated,
  title={Automated speed and lane change decision making using deep reinforcement learning},
  author={Hoel, Carl-Johan and Wolff, Krister and Laine, Leo},
  booktitle={2018 21st International Conference on Intelligent Transportation Systems (ITSC)},
  pages={2148--2155},
  year={2018},
  organization={IEEE}
}

@article{cui2023drivellm,
  title={DriveLLM: Charting the path toward full autonomous driving with large language models},
  author={Cui, Yaodong and Huang, Shucheng and Zhong, Jiaming and Liu, Zhenan and Wang, Yutong and Sun, Chen and Li, Bai and Wang, Xiao and Khajepour, Amir},
  journal={IEEE Transactions on Intelligent Vehicles},
  year={2023},
  publisher={IEEE}
}

@article{duan2024prompting,
  title={Prompting Multi-Modal Tokens to Enhance End-to-End Autonomous Driving Imitation Learning with LLMs},
  author={Duan, Yiqun and Zhang, Qiang and Xu, Renjing},
  journal={arXiv preprint arXiv:2404.04869},
  year={2024}
}

@article{fu2024limsim++,
  title={LimSim++: A Closed-Loop Platform for Deploying Multimodal LLMs in Autonomous Driving},
  author={Fu, Daocheng and Lei, Wenjie and Wen, Licheng and Cai, Pinlong and Mao, Song and Dou, Min and Shi, Botian and Qiao, Yu},
  journal={arXiv preprint arXiv:2402.01246},
  year={2024}
}

@article{rayfeedback,
  title={Feedback-Guided Autonomous Driving},
  author={Ray, Jimuyang Zhang Zanming Huang Arijit and Ohn-Bar, Eshed}
}

@inproceedings{wang2022adept,
  title={ADEPT: A testing platform for simulated autonomous driving},
  author={Wang, Sen and Sheng, Zhuheng and Xu, Jingwei and Chen, Taolue and Zhu, Junjun and Zhang, Shuhui and Yao, Yuan and Ma, Xiaoxing},
  booktitle={Proceedings of the 37th IEEE/ACM International Conference on Automated Software Engineering},
  pages={1--4},
  year={2022}
}

@article{tan2023language,
  title={Language conditioned traffic generation},
  author={Tan, Shuhan and Ivanovic, Boris and Weng, Xinshuo and Pavone, Marco and Kraehenbuehl, Philipp},
  journal={arXiv preprint arXiv:2307.07947},
  year={2023}
}

@article{sun2023reinforcement,
  title={Reinforcement learning in the era of llms: What is essential? what is needed? an rl perspective on rlhf, prompting, and beyond},
  author={Sun, Hao},
  journal={arXiv preprint arXiv:2310.06147},
  year={2023}
}

@article{kirk2023understanding,
  title={Understanding the effects of rlhf on llm generalisation and diversity},
  author={Kirk, Robert and Mediratta, Ishita and Nalmpantis, Christoforos and Luketina, Jelena and Hambro, Eric and Grefenstette, Edward and Raileanu, Roberta},
  journal={arXiv preprint arXiv:2310.06452},
  year={2023}
}

@article{zheng2024balancing,
  title={Balancing Enhancement, Harmlessness, and General Capabilities: Enhancing Conversational LLMs with Direct RLHF},
  author={Zheng, Chen and Sun, Ke and Wu, Hang and Xi, Chenguang and Zhou, Xun},
  journal={arXiv preprint arXiv:2403.02513},
  year={2024}
}

@inproceedings{zhu2023principled,
  title={Principled reinforcement learning with human feedback from pairwise or k-wise comparisons},
  author={Zhu, Banghua and Jordan, Michael and Jiao, Jiantao},
  booktitle={International Conference on Machine Learning},
  pages={43037--43067},
  year={2023},
  organization={PMLR}
}

@book{russell2016artificial,
  title={Artificial intelligence: a modern approach},
  author={Russell, Stuart J and Norvig, Peter},
  year={2016},
  publisher={Pearson}
}

@article{ziegler2019fine,
  title={Fine-tuning language models from human preferences},
  author={Ziegler, Daniel M and Stiennon, Nisan and Wu, Jeffrey and Brown, Tom B and Radford, Alec and Amodei, Dario and Christiano, Paul and Irving, Geoffrey},
  journal={arXiv preprint arXiv:1909.08593},
  year={2019}
}

@article{stiennon2020learning,
  title={Learning to summarize with human feedback},
  author={Stiennon, Nisan and Ouyang, Long and Wu, Jeffrey and Ziegler, Daniel and Lowe, Ryan and Voss, Chelsea and Radford, Alec and Amodei, Dario and Christiano, Paul F},
  journal={Advances in Neural Information Processing Systems},
  volume={33},
  pages={3008--3021},
  year={2020}
}

@online{cosmos-lab,
    title = {COSMOS Lab},
    url = {https://cosmos-lab.org/},
    note = {Accessed: 2024-05-26}
}

@online{cait-datacity,
    title = {DataCity Smart Mobility Testing Ground},
    url = {https://cait.rutgers.edu/datacity/},
    note = {Accessed: 2024-05-26}
}


\end{document}